\documentclass{article}

\usepackage{PRIMEarxiv}

\usepackage[utf8]{inputenc} 
\usepackage[T1]{fontenc}    
\usepackage{hyperref}       
\usepackage{url}            
\usepackage{booktabs}       
\usepackage{amsfonts}       
\usepackage{nicefrac}       
\usepackage{microtype}      
\usepackage{lipsum}
\usepackage{fancyhdr}       
\usepackage{graphicx}       
\graphicspath{{media/}}     
\usepackage{multirow}%
\usepackage{amsmath,amssymb}%
\usepackage{amsthm}%
\usepackage[title]{appendix}%
\usepackage{xcolor}%
\usepackage{textcomp}%
\usepackage{manyfoot}%
\usepackage{booktabs}%
\usepackage{algorithm}%
\usepackage{algorithmicx}%
\usepackage{algpseudocode}%
\usepackage{listings}%
\usepackage{array}
\usepackage{adjustbox}
\usepackage{makecell}
\usepackage{threeparttable}
\usepackage{rotating}
\usepackage{natbib}
\newtheorem{definition}{Definition}%

\title{A Scenario-Driven Cognitive Approach to Next-Generation AI Memory
}

\author{
  Linyue Cai$^{1,3,\dag}$,
  Yuyang Cheng$^{2,3,\dag}$,
  Xiaoding Shao$^{1}$,
  Huiming Wang$^{1}$,
  Yong Zhao$^{1,*}$,
  Wei Zhang$^{3,*}$,
  Kang Li$^{3,*}$ \\
  \\
  $^{1}$School of Computer Science, Sichuan University, Chengdu, China \\
  $^{2}$School of Cyber Science and Engineering, Sichuan University, Chengdu, China \\
  $^{3}$West China Biomedical Big Data Center, Sichuan University West China Hospital, Chengdu, China \\
  \\
  \texttt{\{linyuecai811, yuyangc125, xiaodingshao, huimin.xxtz\}@gmail.com} \\
  \texttt{yong.zhao@scupi.cn, \{zhangwei, likang\}@wchscu.cn} \\
  \\
  $^{\dag}$These authors contributed equally to this work. \\
  $^{*}$Corresponding authors
}

\begin{document}
\maketitle

\begin{abstract}
As artificial intelligence advances toward artificial general intelligence (AGI), the need for robust and human-like memory systems has become increasingly evident. Current memory architectures often suffer from limited adaptability, insufficient multimodal integration, and an inability to support continuous learning. To address these limitations, we propose a scenario-driven methodology that extracts essential functional requirements from representative cognitive scenarios, leading to a unified set of design principles for next-generation AI memory systems. Based on this approach, we introduce the \textbf{COgnitive Layered Memory Architecture (COLMA)}, a novel framework that integrates cognitive scenarios, memory processes, and storage mechanisms into a cohesive design. COLMA provides a structured foundation for developing AI systems capable of lifelong learning and human-like reasoning, thereby contributing to the pragmatic development of AGI.
\end{abstract}


\section{The Human Brain Memory System}
The human brain's memory system consists of three levels: sensory memory, short-term memory, and long-term memory, each corresponding to different brain regions\cite{atkinson1968human,tulving1990priming,squire2009memory}. Sensory memory is processed by the sensory cortex to handle transient information; short-term memory relies on the prefrontal and parietal cortices for temporary storage\cite{sperling1960information}; and long-term memory is consolidated and stored permanently by the hippocampus and neocortex\cite{goldman1995cellular,baddeley2000episodic}\cite{scoville1957loss,mcclelland1995there,dudai2004neurobiology}.
\begin{figure}[H]
    \centering
    \includegraphics[width=1\linewidth]{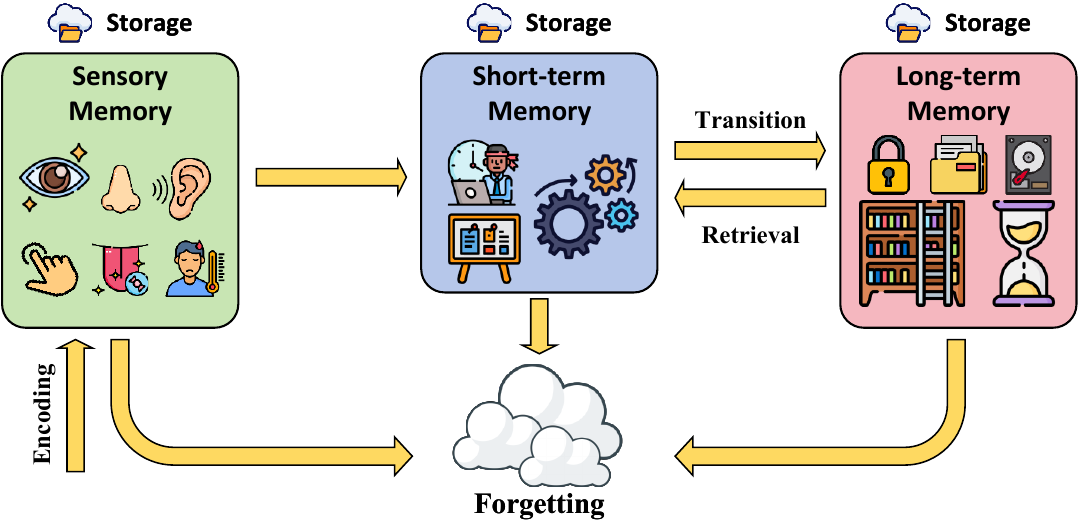}
    \caption{Human Brain's Memory System}
    \label{fig:human-memory}
\end{figure}
The human brain's memory system operates through the coordinated functioning of five key neural mechanisms to achieve efficient information processing, as shown in \autoref{fig:human-memory}. First is the encoding stage, where external information is converted into neural electrical signals by the sensory cortex\cite{logothetis2001neurophysiological,kandel2014molecular,rissman2004measuring}; next is the consolidation process, where the hippocampus converts these signals into stable long-term memory\cite{scoville1957loss,mcclelland1995there,dudai2004neurobiology,frankland2005organization}; followed by the storage phase, where different types of information are categorized and stored in specific brain regions; then comes the retrieval function, where the brain can quickly retrieve memories based on the neural connections established during storage\cite{ratcliff1978theory,bouton1993context}; Finally, the forgetting mechanism intelligently filters and removes redundant information\cite{underwood1957interference,scoville1957loss,ricoeur2004memory}. This sophisticated memory system not only enables long-term storage and rapid retrieval of information but also dynamically optimizes the allocation of cognitive resources\cite{wixted2004psychology}.

In contrast, current AI memory systems remain at a rudimentary level of data storage, lacking systematic design and implementation of key cognitive functions such as collaborative processing and dynamic memory integration. As a result, they are unable to authentically replicate the complex operational mechanisms of human brain memory. This gap underscores the urgent need to draw deeper inspiration from the working principles of biological memory, in order to guide the design and evolution of next-generation memory architectures.

\section{Overall Limitations of Existing AI Memory}
\begin{figure}
\centering
\includegraphics[width=1\linewidth]{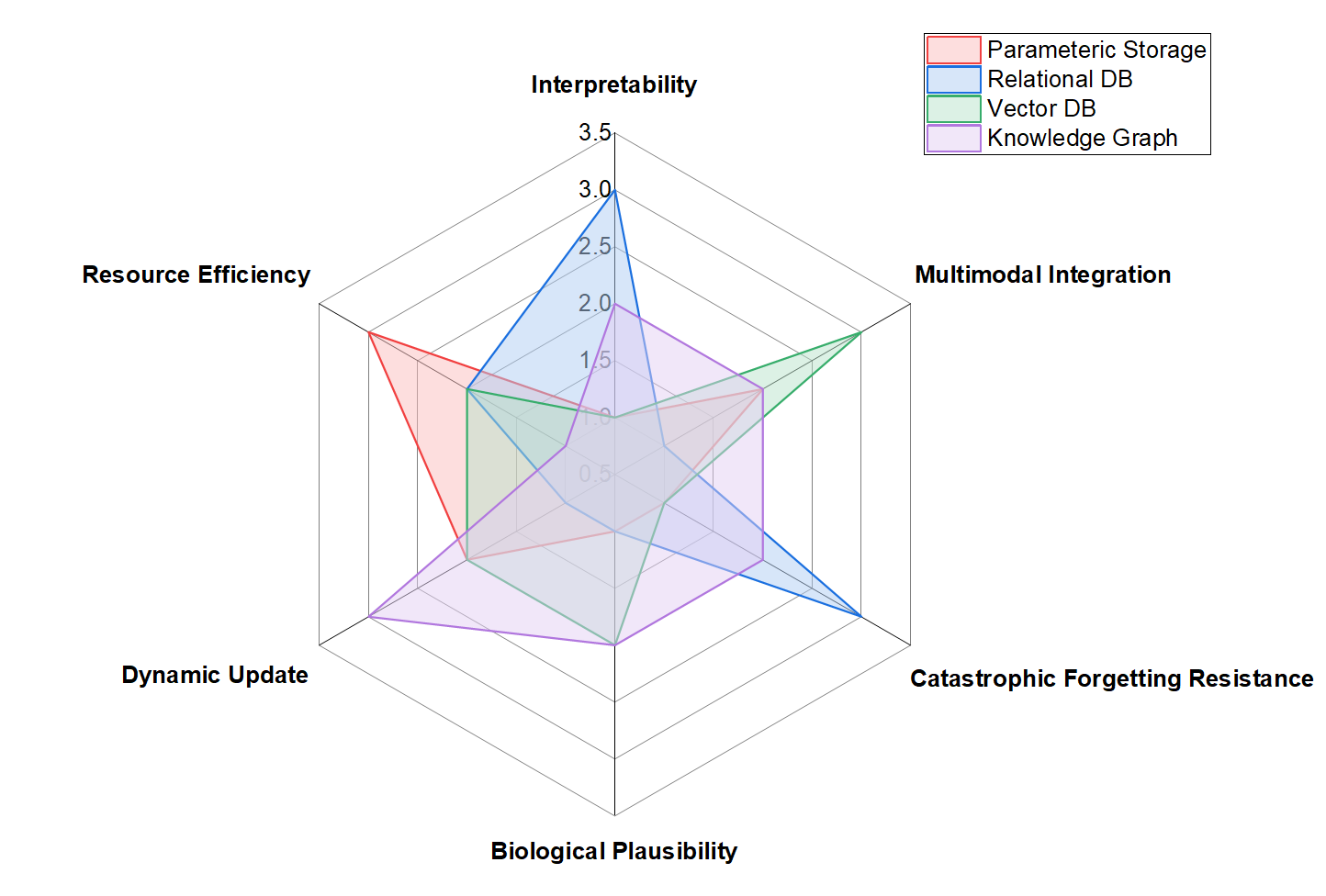}
\caption{AI Memory Systems: Six-Dimension Evaluation}
\label{fig:ai_memory_evaluation}
\end{figure}

Current mainstream artificial intelligence systems primarily rely on four memory storage paradigms: parameterized storage in language models\cite{bengio2013representation,10.1007/978-3-642-42051-1_16,zhao2023survey,gao2023retrieval}, relational databases\cite{maier1983theory,halpin2010information,battaglia2018relational}, vector databases\cite{sukhbaatar2015end,baltruvsaitis2018multimodal, johnson2019billion}, and knowledge graphs based on triples\cite{10.1007/978-3-319-93417-4_38,hogan2021knowledge}. While these architectures have enabled significant progress in AI capabilities, they exhibit critical limitations when assessed against the dynamic, adaptive, and integrative nature of human memory\cite{robins1995catastrophic, schacter2007remembering,hassabis2017neuroscience,tulving1972episodic}. The six-dimensional evaluation in \autoref{fig:ai_memory_evaluation} provides a comprehensive visual summary of these limitations, revealing that no single existing approach excels across all criteria—underscoring the need for more holistic memory architectures.

Human cognition demonstrates an innate ability to dynamically update, integrate, and prioritize information based on relevance and experience\cite{thelen1994dynamic,newell1994unified,baddeley2000episodic}. As clearly depicted in \autoref{fig:ai_memory_evaluation}, existing AI memory systems struggle particularly in the dimensions of dynamic update capability and catastrophic forgetting resistance. Parameterized storage in language models suffers from catastrophic forgetting\cite{mccloskey1989catastrophic,robins1995catastrophic,kirkpatrick2017overcoming,pmlr-v80-serra18a}, as evidenced by its poor performance on forgetting resistance in the evaluation, while structured databases require manual intervention for updates, showing limited dynamic update capacity that makes them ill-suited for scenarios requiring continuous learning. The human brain, by comparison, refines knowledge progressively without erasure, balancing stability with plasticity—a capability current AI architectures lack\cite{hebb2005organization,bliss1973long,kandel2001molecular}.

The dimension of multi-modal integration in \autoref{fig:ai_memory_evaluation} further reveals a critical gap in memory integration. Human cognition effortlessly combines sensory inputs, emotional context, and abstract knowledge into a unified representation\cite{damasio1989brain,newell1994unified,damasio1999feeling,shams2008benefits,buckner2007self,binder2011neurobiology}. For instance, recalling a familiar face involves not just visual data but associated emotions, conversations, and spatial context. However, as the evaluation shows, existing AI systems treat memory modules as isolated components—vector databases store perceptual data separately from symbolic knowledge graphs, preventing the kind of cross-modal reasoning that defines human intelligence\cite{baltruvsaitis2018multimodal}. This fragmentation limits applications in areas like social robotics or interactive AI, where real-world understanding depends on synthesizing diverse information streams.

The resource efficiency dimension in the evaluation highlights another key differentiator between biological and artificial memory. The brain dynamically strengthens or prunes memories based on importance and frequency, optimizing cognitive load\cite{tononi2014sleep}. AI systems, in contrast, as reflected in their medium to low scores on resource efficiency, rely on static storage strategies, retaining irrelevant data while struggling to prioritize critical information\cite{rolnick2019experience}. This inefficiency becomes apparent in real-time decision-making scenarios, such as autonomous navigation, where persistent safety rules must override transient sensory noise—a balance current architectures cannot achieve without explicit programming.

Finally, the interpretability dimension in \autoref{fig:ai_memory_evaluation} exposes a fundamental challenge shared by most AI memory systems. Human memory supports introspection—we can often explain why we recall certain details and how they relate to broader knowledge. AI memory representations, whether distributed embeddings or opaque model parameters, as visually confirmed by their low interpretability scores, lack such transparency\cite{doshi2017towards,samek2021explaining}. This limits their applicability in high-stakes domains like healthcare or law, where traceable reasoning is essential.

The comprehensive assessment provided by \autoref{fig:ai_memory_evaluation} clearly indicates that incremental improvements to existing paradigms will not suffice. Instead, future AI memory systems must draw inspiration from neuroscience, incorporating mechanisms for dynamic consolidation, cross-modal association, and resource-aware storage\cite{hassabis2017neuroscience}. The uniformly low scores in biological plausibility across all paradigms further reinforce this conclusion. By grounding design in human cognitive scenarios—from lifelong learning to explainable decision-making—we can move beyond rigid, isolated memory architectures toward systems that truly emulate the flexibility and richness of biological memory. In the following sections, we will examine four real-world cognitive scenarios to concretely analyze the key limitations of current AI memory systems. 

\section{Scenario Driven Cognitive Behavior}

\subsection*{S1: Toxic Mushroom Identification - A Cognitive Process Demonstration}

\begin{definition}[Toxic Mushroom Identification Process]
When you come across a mushroom in the wild, you first carefully examine its color and shape, then gently touch it to feel its texture. If still uncertain about its toxicity, you promptly take out your phone to photograph and identify it. When the app displays a "highly toxic" warning, you quickly step back to avoid it while firmly committing its characteristics to memory.
\end{definition}

\begin{figure}[H]
    \centering
    \includegraphics[width=1\linewidth]{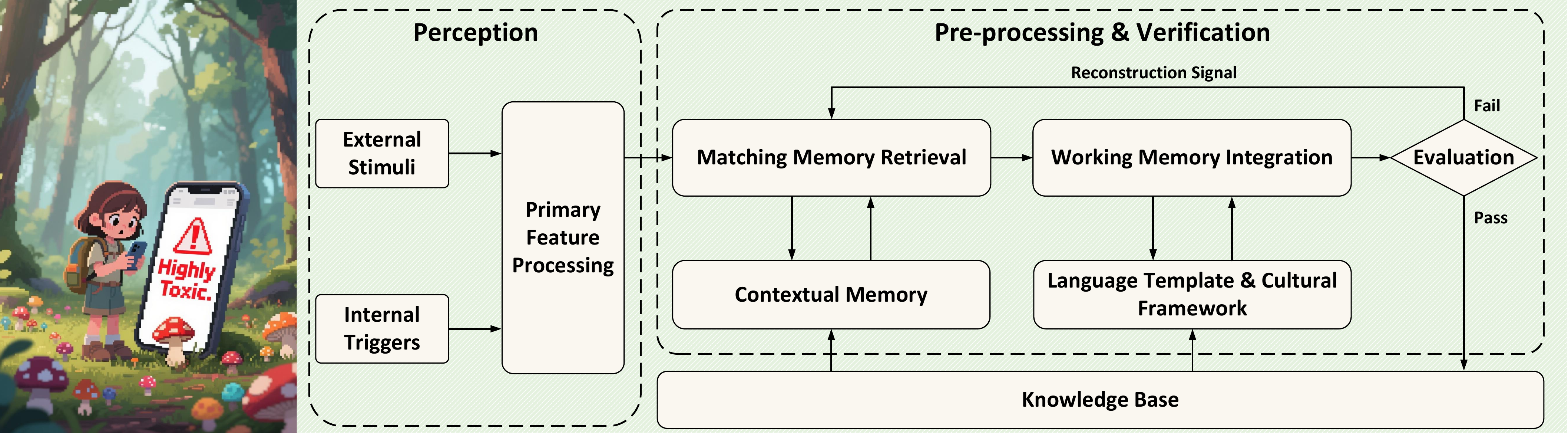}
    \caption{Cognitive Process of Toxic Mushroom Identification}
    \label{fig:mushroom_identification}
\end{figure}
When identifying mushrooms, the brain works through association and prediction. Visual, tactile, and olfactory information are rapidly associated with stored mushroom characteristics in memory, enabling danger prediction and swift judgment\cite{barsalou1999perceptual}. The cognitive workflow depicted in \autoref{fig:mushroom_identification} illustrates this integrated processing mechanism.

Modern cognitive informatics provides a digital and coherent framework for explaining human cognitive processes \citep{sciencedirect2024}. Natural intelligence evolves by repeatedly applying simple navigation techniques in different coordinates millions of times \citep{li2024organizational}. This phenomenon is as described by the ecological perception theory proposed by \citep{Gibson1966}, where the immediate characteristics of environmental stimuli directly form initial perceptual representations. This pre-conscious processing provides a necessary buffer platform for subsequent semantic analysis. At this point, the perceptual information is still in an unprocessed state, similar to the preparatory stage set by the nervous system for subsequent cognitive activities. 
During the stage of deep information processing, the brain activates the hippocampus - medial temporal lobe memory circuit to extract the long-term stored semantic knowledge \citep{squire2004memory,phelps2004human,tanaka2025key}. This process not only involves the routine retrieval of foreign language vocabulary, but also includes the topological activation of related cultural imagery - ranging from typical phonetic rhythm features to social interaction conversation frameworks \citep{Squire2004,hagoort2005broca,federmeier2007thinking}. Hof et al.\cite{Hof2003} confirmed through functional magnetic resonance imaging studies that the oscillations of the temporal cortex are significantly correlated with the associative efficiency of cross-cultural semantics, and this synchronized neural activity provides the necessary bioelectric basis for real-time language understanding \citep{Hof2003}. 
The cognitive resource allocation dominated by the prefrontal cortex constitutes the final link of the cognitive loop. This region maintains the focus of attention and information storage, thereby dynamically matching the immediate perceptual signals with the long-term cultural semantics \citep{rolls2008memory}. As the joint experiment by Smith et al.\cite{Smith2006} revealed, when the subjects needed to process foreign accents and facial expressions simultaneously, the BOLD signal intensity of the dorsolateral prefrontal cortex showed a characteristic fluctuation pattern \citep{Smith2006}. The efficiency of integrating multi-dimensional information is essentially derived from the temporal and spatial encoding mechanism of the working memory system for limited cognitive resources \citep{Baddeley1992}.

\subsection*{S2: Daily Recall - A Dynamic Memory Reconstruction Process Demonstration}

\begin{definition}[Daily Recall Process]
When trying to recall what you did on the 2nd of last month, your brain first determines what day of the week it was - using this to retrieve regular weekly routines (e.g., Monday meetings). If it was a weekend, you'd recall any special plans. If still unclear, you might work backward from activities on the 1st or consult external cues like photos in the phone and chat histories to trigger memories.
\end{definition}
\begin{figure}[H]
    \centering
    \includegraphics[width=1\linewidth]{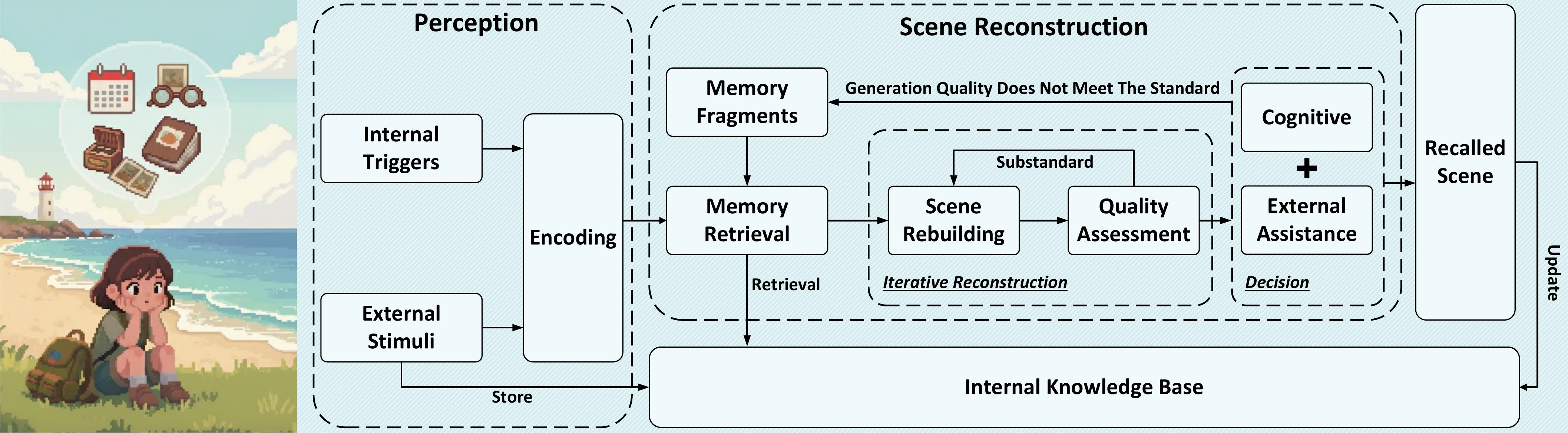}
    \caption{Dynamic Memory Reconstruction Process of Daily Recall}
    \label{fig:memory_reconstruction}
\end{figure}

When recalling events from a specific date, the brain primarily employs two core memory functions: Recall and Association. The Recall mechanism retrieves routine activity memories through temporal frameworks, while the Association function dynamically integrates date information with external environmental cues to jointly accomplish memory reconstruction. This complex reconstruction process is visualized in \autoref{fig:memory_reconstruction}.

Typically, it takes thirty seconds to a minute of focused reflection, during which the hippocampus and associated cortical areas collaborate\cite{eichenbaum2017memory, kwon2025coordinated,raud2023hippocampal,duff2020semantic}. External cues often support the otherwise vague mental imagery\cite{tulving1973encoding}.

Initially, the brain encodes exteroceptive inputs (e.g., sounds, smells, visuals) and interoceptive cues (e.g., keywords, emotions) into fragments of short-term memory held in working memory. These fragments activate long-term memory networks, retrieving relevant traces and integrating them into an initial multimodal scaffold\cite{barsalou1999perceptual,ranganath2010binding}.

The brain then enters an iterative reconstruction phase, repeatedly assembling and refining these fragments\cite{shi2024dynamic,addis2018episodic}. Each cycle is evaluated for completeness, coherence, and affective congruence. Fragments that fall short are revised and reprocessed in subsequent iterations. This resembles a computational feedback loop, gradually converging on a scene representation that approximates the original memory.

Despite such refinement, recall can still distort reality—objects may be misremembered, or contextual features may shift due to imagination or interference. Each act of recall also updates long-term memory, reinforcing accurate details or consolidating altered ones\cite{alberini2013memory}.

Understanding this dynamic reconstruction mechanism shown in \autoref{fig:memory_reconstruction} provides key insights for designing memory systems in artificial intelligence\cite{kim2023transformer,whittington2021relating}—highlighting the roles of multimodal fusion, iterative feedback, and synaptic plasticity\cite{hassabis2017neuroscience,spens2024generative,zhou2023self,kemker2017fearnet}. It also informs neuroscience-based approaches to treating memory disorders and developing cognitive enhancement technologies\cite{du2025rethinking}.

\subsection*{S3: Mathematical Problem-Solving - A Reasoning Process Demonstration}

\begin{definition}[Mathematical Problem-Solving Process]
When presented with a problem, you quickly identify the question type and objectives, analyze the given conditions to determine the solution approach, simultaneously generate multiple solution pathways in your mind while retrieving relevant formulas, verify their validity before proceeding with the optimal derivation, record the final answer, and ultimately review the entire process to consolidate insights and refine your problem-solving strategies.
\end{definition}

\begin{figure}[H]
    \centering
    \includegraphics[width=\linewidth]{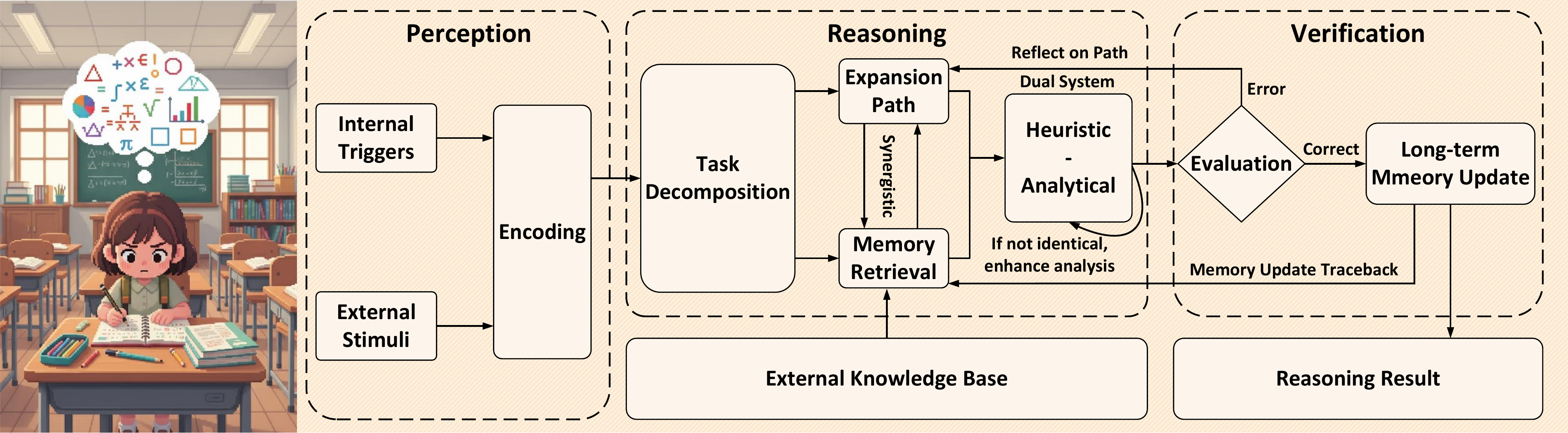}
    \caption{Reasoning Process of Mathematical Problem-Solving}
    \label{fig:reasoning_process}
\end{figure}

The mathematical reasoning process relies on Reasoning for and Reflection. When solving problems, the Reasoning function enables systematic manipulation of mathematical concepts through rule-based operations, while Reflection facilitates critical evaluation of both successful and unsuccessful solution strategies, leading to improved future performance. The complete reasoning cycle is systematically depicted in ~\autoref{fig:reasoning_process}.

In a complete reasoning cycle, individuals begin by perceiving and encoding either external stimuli or internally initiated task-related information. This information is temporarily maintained within working memory, primarily supported by the dorsolateral prefrontal cortex (DLPFC), and is subsequently decomposed into subgoals for further processing\cite{baddeley2012working,Miller2001-hm}. Next, the parietal-frontal network—working in conjunction with the medial temporal lobe—facilitates the rapid retrieval of relevant long-term memories\cite{Cabeza2000-jj}. This network also underpins the expansion of associative links and the integration of relational information through divergent activation along hypothesized “reasoning chain”\cite{christoff2001rostrolateral, Jung2007-kd}. We would suggest that this part of the process could be complemented by the use of an external knowledge base. The retrieved episodic or semantic memory elements function as “knowledge cues” that are concurrently processed by a dual-system architecture. Specifically, the temporal-frontal association system generates intuitive heuristics, whereas the frontal-parietal executive network engages in rule-based, deductive or inductive inference\cite{prado2011brain,Goel2004-rl}. These two systems operate in parallel and their outputs are subject to conflict monitoring by the anterior cingulate cortex (ACC); if an inconsistency between the intuition and the rule is detected, the ACC calls back to the frontal lobe to enhance depth analysis\cite{botvinick2001conflict,Kerns2004-ja}. The preliminary outcome is subsequently routed to an evaluation module. If the output is assessed as correct, consolidation mechanisms are triggered, promoting the encoding of new rules into long-term memory\cite{dudai2004neurobiology}. Conversely, if the evaluation considers the outcome unsatisfactory, the process is recursively directed back to the retrieval stage. There, adjustments to the knowledge structure and problem-solving strategy are made. This iterative loop—comprising stages of perception, decomposition, retrieval, reasoning, verification, and reinforcement—ultimately yields validated reasoning outcomes and forms structured cognitive traces that support future recall and generalization.

\subsection*{S4: Historical Knowledge Updating - A Memory Updating Process Demonstration}

\begin{definition}[Historical Knowledge Updating Process]
When reading a newly published historical biography, you encounter a description of Napoleon that conflicts with your high school history lessons. The biography suggests that Napoleon's decisions during the Battle of Waterloo were primarily influenced by intelligence failures, rather than the simplistic "defeat caused by stubbornness" explanation stored in your memory.
\end{definition}
\begin{figure}[H]
    \centering
    \includegraphics[width=1\linewidth]{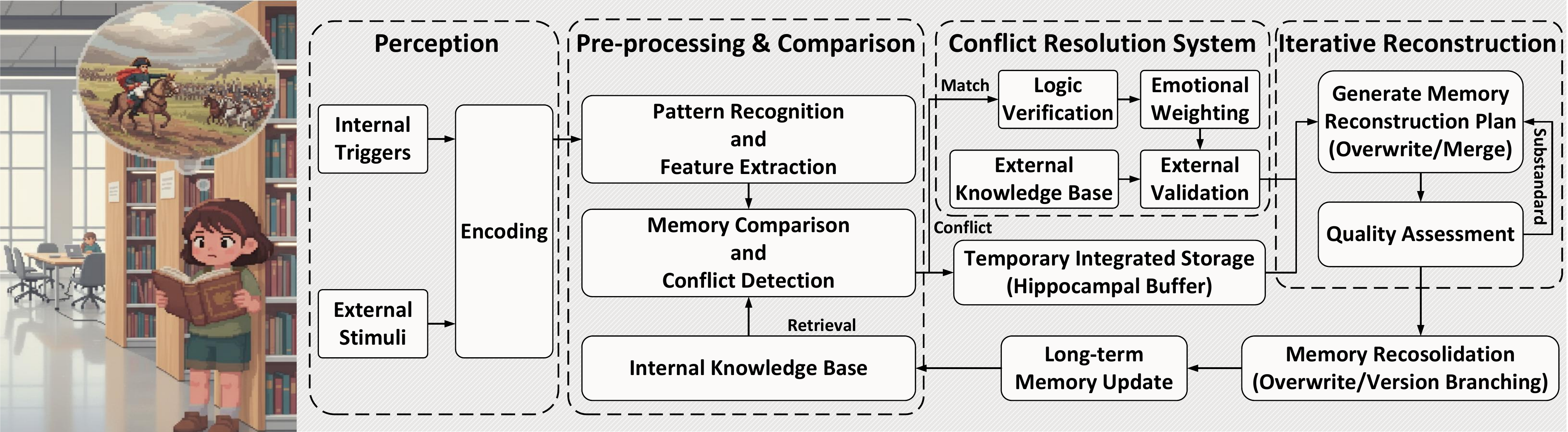}
    \caption{Memory Updating Process of Historical Knowledge Updating}
    \label{fig:memory_updating}
\end{figure}

The memory system employs both recall and continual learning when processing new historical information. Recall retrieves prior knowledge while continual learning integrates new facts, enabling adaptive understanding of historical events. This sophisticated updating mechanism is detailed in \autoref{fig:memory_updating}.

When you are reading a new book about history, a historical event mentioned in the book may be something you already know about. The new information enters the sensory cortex of the brain for encoding, while the memory you previously stored about this event is activated. At this point, the association cortex region of the brain compares and integrates the details mentioned in the new book with your existing memory. If the new information conflicts with the old memory, the brain acts like an advanced scanner to reassess and update the memory.

First, external stimuli and internal triggers are encoded through the sensory cortex, with preliminary pattern recognition and feature extraction occurring in the association cortex regions of the temporal and parietal lobes\cite{squire1991medial,hubel1962receptive,ungerleider1982two}. This process is akin to an advanced scanner breaking down and categorizing information. Next, the information enters the memory preprocessing and comparison phase. The hippocampus, as the central relay station, retrieves relevant existing knowledge from the long-term memory storage area for enhancement and comparison\cite{bliss1993synaptic,kumaran2016learning,buzsaki2015hippocampal,squire2011cognitive,moser2017spatial}. This process differs fundamentally from simple key-value retrieval, resembling more closely a knowledge-graph-style exploration that expands outward from core nodes along semantic associative networks\cite{cai2024bringing,cai2025practices}. If the new information highly matches existing memories, the prefrontal cortex activates the default mode network, temporarily integrating and storing it in the hippocampus buffer\cite{dudai2015consolidation,shohamy2010dopamine,nader2009single,rossato2009dopamine}. If conflicts are detected, the brain initiates a deep quality check: the prefrontal cortex performs logical verification, the amygdala assigns emotional weight to adjust priorities, and external databases are consulted for cross-validation\cite{botvinick2001conflict,cohen2013midfrontal,shenhav2013expected}. Subsequently, the information enters the core validation loop. The system generates multiple reconstruction schemes and conducts multiple rounds of quality assessments based on consistency and completeness. Memories that fail to meet the standards are returned for reprocessing until they pass the quality check. Finally, memories that meet the criteria enter long-term storage through a reconsolidation mechanism: either by being integrated into the existing memory network through long-term potentiation (LTP) mechanisms\cite{bliss1973long}, or by forming new independent memory traces\cite{tonegawa2015memory}.

This process reveals the dynamic nature of memory updating—it is not a passive storage process but an active reconstruction process involving collaboration among multiple brain systems, ensuring the validity and adaptability of information. Based on observations of human memory updating and related research, we define memory updating as the process by which an intelligent agent performs memory retrieval, comparison, verification, and reconsolidation, and have summarized the memory updating mechanism as shown in \autoref{fig:memory_updating}.

\section{A Scenario-Driven Cognitive Capability Framework for AI Memory Systems}

Traditional memory mechanisms are often reduced to "data warehouses" for basic storage and retrieval. This paradigm requires scenario-driven re-evaluation: deducing essential capabilities from target operational scenarios. Next-generation memory must become core infrastructure for AGI cognition, reasoning, and autonomous evolution, possessing five critical capabilities:

\begin{enumerate}
    \item \textbf{Reasoning:} Logical/causal inference from memory.
    \item \textbf{Recall:} Precise, efficient information retrieval.
    \item \textbf{Association:} Cross-domain/cross-modal knowledge linking.
    \item \textbf{Prediction:} Anticipating future states based on patterns.
    \item \textbf{Reflection:} Self-evaluation and correction.
    \item \textbf{Continual Learning:} Integrating new knowledge while preserving stability.
\end{enumerate}

In evaluation frameworks, conventional accuracy-centric metrics fail to capture memory systems' real-world utility. Addressing AGI’s requirements necessitates transcending traditional paradigms through a contextualized framework that assesses operational efficacy via multidimensional dimensions: cognitive capacity, adaptive agility, and evolutionary scalability.

To address this, we construct a comparative framework encompassing twelve multifaceted dimensions—including multimodal support, similarity retrieval, and indexing mechanisms—systematically characterizing critical capabilities of mainstream memory storage technologies and hybrid architectures. This framework establishes a comprehensive, future-oriented evaluation paradigm by visualizing capability coverage across existing solutions, offering actionable insights for next-generation memory system design and optimization.The results are shown in \autoref{compare}.

\begin{sidewaystable*}
\centering
\footnotesize
\setlength{\tabcolsep}{3.5pt}
\begin{threeparttable}
\caption{Comparison of Memory System Capabilities Across 12 Architectural Dimensions}
\label{compare}
\begin{tabular}{@{}l|cccccccccccc@{}}
\toprule
System Combination & 
\rotatebox{45}{Multi-modal} & 
\rotatebox{45}{Similarity} & 
\rotatebox{45}{Indexing} & 
\rotatebox{45}{Sync} & 
\rotatebox{45}{Entity Model} & 
\rotatebox{45}{Time Series} & 
\rotatebox{45}{Versioning} & 
\rotatebox{45}{Distributed} & 
\rotatebox{45}{Linking} & 
\rotatebox{45}{Compression} & 
\rotatebox{45}{Online Update} & 
\rotatebox{45}{Reasoning} \\
\midrule
MLP                & Y & N & P & P & N & P & P & P & N & Y & Y & N \\
KG                 & P & N & P & N & Y & N & N & N & Y & N & Y & Y \\
KV Store           & N & N & Y & P & N & Y & P & Y & N & P & Y & N \\
VectorDB           & P & Y & N & P & N & Y & N & P & N & Y & Y & N \\
MLP + VectorDB     & Y & Y & P & P & N & Y & P & Y & P & Y & Y & N \\
KG + MLP           & Y & P & P & P & Y & P & P & P & Y & Y & Y & Y \\
KG + VectorDB      & Y & Y & P & N & Y & Y & N & Y & Y & Y & Y & Y \\
KG + VectorDB + MLP& Y & Y & Y & P & Y & Y & P & Y & Y & Y & Y & Y \\
Relational DB      & N & N & P & P & Y & N & P & N & N & N & Y & N \\
MongoDB            & P & N & Y & P & N & N & P & Y & N & P & Y & N \\
Cassandra          & Y & N & P & Y & N & Y & N & Y & N & P & Y & N \\
VectorDB + MongoDB & Y & Y & P & P & N & Y & P & Y & P & Y & Y & N \\
VectorDB + Cassandra&P& Y & P & Y & N & Y & N & Y & N & P & Y & N \\
\bottomrule
\end{tabular}
\begin{tablenotes}
\item \textbf{Legend:} Y = Fully supported, N = Not supported, P = Partially supported
\end{tablenotes}
\end{threeparttable}
\end{sidewaystable*}

\section{Cognition Layered Memory Architecture}

\begin{figure}
    \centering
    \includegraphics[width=\linewidth]{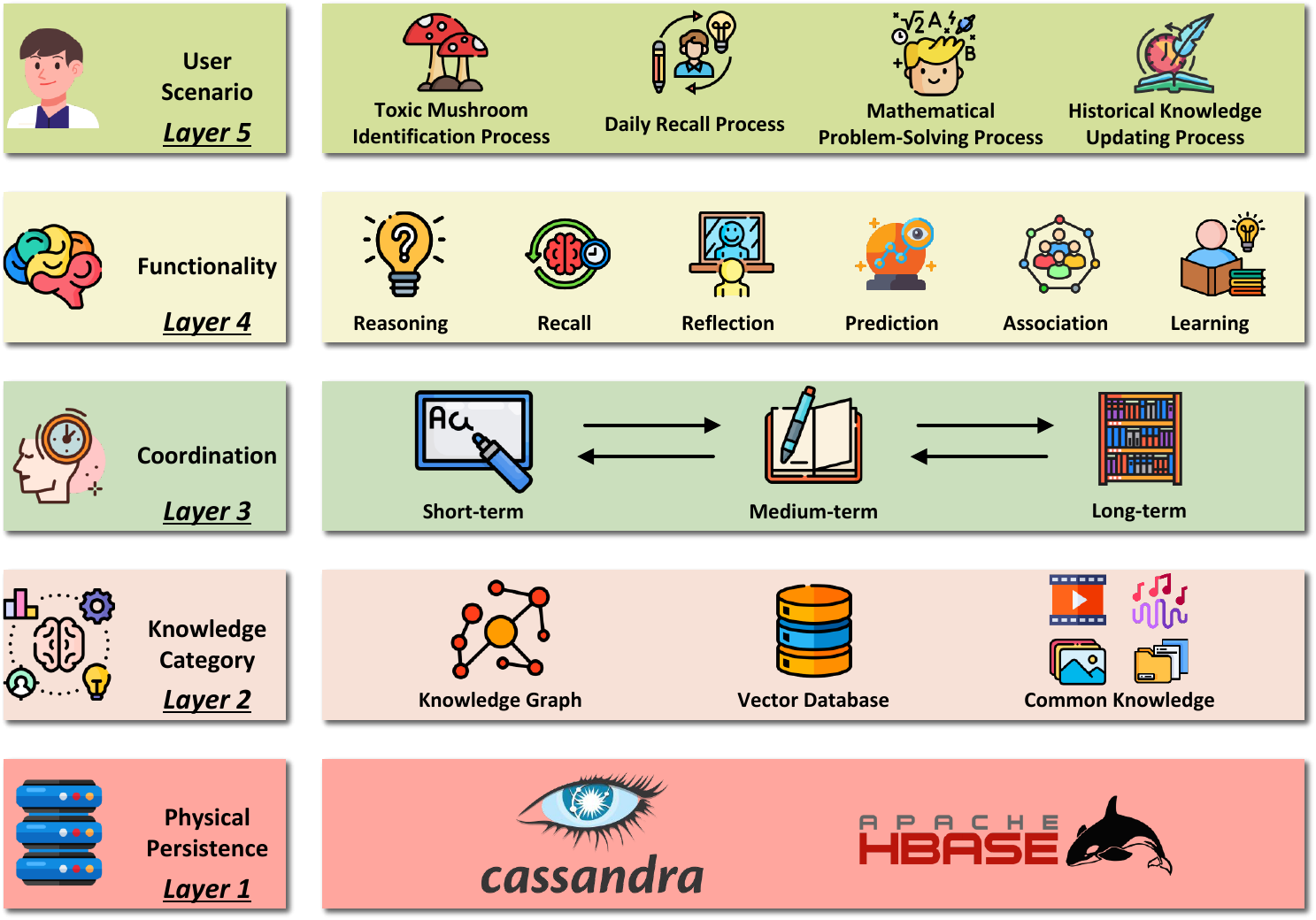}
    \caption{Framework of COgnitive Layered Memory Architecture}
    \label{fig:COLMA}
\end{figure}

As discussed above, current artificial intelligence memory systems exhibit significant deficiencies in dynamic adaptability, multimodal integration, resource efficiency, and interpretability, rendering them inadequate for supporting complex cognitive tasks in real-world scenarios. First, static storage architectures fail to accommodate dynamic cognitive demands: parametric storage suffers from catastrophic forgetting, while traditional databases rely heavily on manual updates. Second, multimodal information is often stored in isolation, resulting in a lack of cross-modal association and integration capabilities. Third, resource allocation mechanisms are rigid, unable to dynamically reinforce critical information or discard redundant memories in a manner akin to the human brain. Finally, memory representations lack interpretability: although embedding vectors support retrieval, they offer limited traceability of the reasoning process. These shortcomings are particularly pronounced in scenarios involving continual learning, complex reasoning, and real-time decision-making, severely constraining the development of AI cognitive abilities.

Analysis of scenarios such as mushroom recognition, everyday memory recall, mathematical problem solving, and historical knowledge updating reveals that the core advantages of human memory lie in its hierarchical coordination mechanisms (e.g., dynamic interactions between sensory and long-term memory), cross-modal association capabilities (e.g., seamless integration of visual and semantic information), and stability and adaptability in continual learning. Furthermore, through a comparative analysis of twelve storage architectures (\autoref{compare}), we find that Cassandra-based architectures exhibit distinct advantages in distributed scalability, multimodal support, online updating, and temporal control. Its flexible columnar storage structure and high-throughput write capability provide a solid physical foundation for constructing hierarchical memory systems.

Based on the above insights, we propose a novel hierarchical artificial intelligence memory architecture—\textbf{\emph{COgnitive Layered Memory Architecture (COLMA)}}—which flexibly leverages either Cassandra\cite{lakshman2010cassandra} or HBase\cite{taylor2010overview} (functionally similar alternatives) as its underlying distributed storage layer.  COLMA is designed to realize the next-generation AI memory system, endowed with dynamic adaptability, cross-modal integration capabilities, and continuous evolvability. Its hierarchical structure is illustrated in \autoref{fig:COLMA}.

The COLMA is organized into five levels, arranged from bottom to top as the Physical Persistence Layer, Knowledge Category Layer, Coordination Layer, Functionality Layer, and User Scenario Layer. COLMA employs a hierarchical design paradigm, organically integrating cognitive scenarios, memory functionalities, and underlying storage to construct a dynamically coordinated and unified system.

At the \emph{User Scenario Layer}, the system flexibly supports a wide range of cognitive and reasoning tasks, achieving a deep coupling between application requirements and memory operations. The \emph{Functionality Layer} integrates core AI capabilities such as reasoning, recall, and association, endowing the system with human-like complex knowledge processing abilities. At the \emph{Coordination Layer}, dynamic collaboration among long-, medium-, and short-term memories simulates the interaction between the hippocampus and neocortex in biological memory, enabling efficient information integration and optimized allocation. The \emph{Knowledge Category Layer} fuses knowledge graphs, vector databases, and common knowledge  to construct multimodal knowledge representations, while leveraging Cassandra’s distributed features for efficient management and retrieval. At the bottom, the \emph{Physical Persistence Layer} relies on Cassandra’s high-performance storage capabilities to ensure reliable data persistence and rapid access.

Overall, COLMA is not limited to a specific technical implementation but serves as a theoretical framework for next-generation AI memory systems. Its core concept lies in combining Cassandra’s elastic scalability and high-performance storage mechanisms with heuristic principles inspired by cognitive science, thereby constructing an intelligent memory system endowed with adaptability, evolvability, and scenario-aware capabilities.

To quantitatively assess the advantages of our proposed COLMA framework, we conducted a systematic evaluation against six prominent memory architectures for AI systems: A-Mem \cite{xu2025mem}, MemO \cite{chhikara2025mem0}, MemOg \cite{chhikara2025mem0}, MEM1 \cite{zhou2025mem1}, MIRIX \cite{wang2025mirix}, and Mem$^p$ \cite{fang2025memp}. The evaluation focused on ten critical dimensions essential for next-generation AI memory systems:

\begin{enumerate}
    \item \textbf{Dynamic Update}: Ability to adaptively modify stored knowledge.
    \item \textbf{Indexing}: Efficiency in organizing and retrieving information.
    \item \textbf{Multimodal Integration}: Support for diverse data types and modalities.
    \item \textbf{Heterogeneous Representation}: Ability to unify diverse data types and formats.
    \item \textbf{Interpretability}: Transparency and explainability of memory operations.
    \item \textbf{Biological Plausibility}: Alignment with human memory mechanisms.
    \item \textbf{Distributed Scalability}: Performance in distributed computing environments.
    \item \textbf{Time Series Handling}: Effectiveness in processing temporal sequences.
    \item \textbf{Associative Reasoning}: Capacity for connection-based inference.
    \item \textbf{User Permission}: Granular access control and data isolation based on user roles or identities to ensure security, privacy, and collaborative integrity.
\end{enumerate}

Each dimension was rated on a three-star scale ($\star$ = Basic, $\star\star$ = Good, $\star\star\star$ = Excellent) based on comprehensive analysis of each architecture's capabilities and limitations. The results, presented in \autoref{tab:memory_evaluation}, demonstrate COLMA's superior performance across all evaluated dimensions.

\newcommand{\one}{$\star$}
\newcommand{\two}{$\star\star$}
\newcommand{\three}{$\star\star\star$}

\begin{table}[ht]
    \centering
    \setlength{\tabcolsep}{4pt}
    \caption{Comparative Evaluation of Memory Architectures}
    \label{tab:memory_evaluation}
    \begin{tabular}{l *{7}{c}}
    \toprule
    \textbf{Evaluation Dimension} 
    & \makecell{\textbf{COLMA}\\\textbf{(Ours)}} 
    & \textbf{A-Mem} 
    & \textbf{MemO} 
    & \textbf{MemOg} 
    & \textbf{MEM1} 
    & \textbf{MIRIX} 
    & \textbf{Mem$^p$} \\
    \midrule
    Dynamic Update                  & \three & \three & \three & \three & \three & \three & \three \\
    Indexing                        & \three & \three & \three & \three & \two   & \three & \two   \\
    Multimodal Integration          & \three & \one   & \one   & \one   & \one   & \two   & \one   \\
    Heterogeneous Representation    & \three & \two   & \one   & \one   & \one   & \three & \two \\
    Interpretability                & \three & \two   & \two   & \three & \one   & \two   & \two   \\
    Biological Plausibility         & \three & \two   & \two   & \two   & \one   & \one   & \one   \\
    Distributed Scalability         & \three & \two   & \two   & \two   & \one   & \three & \one   \\
    Time Series Handling            & \three & \two   & \two   & \two   & \two   & \three & \two   \\
    Associative Reasoning           & \three & \three & \two   & \three & \two   & \two   & \two   \\
    User Permission                 & \three & \one   & \one   & \one   & \one   & \two   & \one   \\
    \midrule
    \textbf{Overall Score} & \textbf{30/30} & \textbf{21/30} & \textbf{19/30} 
                           & \textbf{21/30} & \textbf{15/30} & \textbf{24/30} & \textbf{17/30} \\
    \bottomrule
    \end{tabular}

    \vspace{2mm}
    \raggedright \small 
    \textbf{Rating Scale:} \one = Basic, \two = Good, \three = Excellent. \\
    \textbf{Scoring:} Each \one = 1 point, maximum 3 points per dimension.
\end{table}

The evaluation reveals several key insights. First, COLMA achieves perfect scores across all dimensions—including heterogeneous representation and user permission partitioning—demonstrating its comprehensive capabilities as a unified memory architecture. This exceptional performance stems from its layered design that integrates Cassandra's distributed storage, multimodal knowledge representation, cognitive-inspired coordination mechanisms, and a flexible encoding framework to unify diverse data types (text, structured tables, multimodal fragments) while preserving their intrinsic properties. Second, while several systems (A-Mem, MemO, MemOg) excel in dynamic updating and indexing capabilities, they uniformly struggle with multimodal integration and heterogeneous representation—critical limitations that COLMA overcomes through its knowledge category layer: this layer not only seamlessly combines knowledge graphs, vector databases, and common knowledge bases but also enables coherent association of disparate data formats, avoiding the siloing of heterogeneous information seen in comparative systems. Third, COLMA stands out in biological plausibility, scoring significantly higher than all comparative systems. This advantage reflects its design inspiration from human memory mechanisms, including hippocampal-neocortical interactions, dynamic memory consolidation processes observed in biological systems, and the human brain’s innate ability to integrate varied types of experiences into unified memory representations—paralleling COLMA’s strength in heterogeneous data unification.

The comparative analysis confirms that COLMA represents a significant advancement over existing memory architectures, offering a more holistic solution that addresses the multifaceted requirements of next-generation AI systems. Crucially, in response to the evolution of artificial intelligence toward collaborative intelligence, COLMA introduces a novel memory paradigm for Multi-Agent Systems (MAS)\cite{cheng2025hawk, wu2023autogen,wang2024agent,duan2024exploration,li2023camel}, overcoming the isolation and staticity inherent in traditional memory modules. Its hierarchical, interpretable, and adaptive architecture enables unified and traceable knowledge sharing across agents, driving a fundamental shift from behavioral coordination to cognitive collaboration—ultimately enabling the emergence of collective intelligence.

\section{Conclusion}
This paper identifies critical limitations of current AI memory systems—in adaptability, multimodal integration, resource efficiency, and interpretability—and draws inspiration from human cognitive scenarios to extract principles of coordination and continual learning. Building on these insights, we introduce the COgnition Layered Memory Architecture (COLMA), which integrates Cassandra’s scalable persistence with heuristic principles from cognitive science. COLMA offers a unified, scenario-driven framework that repositions memory not as static storage, but as an adaptive, multimodal, and evolvable substrate for intelligence. Looking ahead, COLMA will be deployed and validated in real-world domains such as healthcare, finance, and scientific research, where its practical utility will be continuously demonstrated—and through which we will iteratively explore and deepen our understanding of AI cognition in context. In summary, COLMA provides a scalable and cognitively inspired foundation for next-generation AI cognition, and we envision it as a cornerstone toward the realization of artificial general intelligence.

\section*{Acknowledgements}

This study was supported by the Interdisciplinary Crossing and Integration of Medicine and Engineering for Talent Training Fund, West China Hospital, Sichuan University;the 1·3·5 project for disciplines of excellence, West China Hospital, Sichuan University(ZYYC21004); the National Natural Science Foundation of China (NSFC) under Grant [No.62177007].

\bibliographystyle{unsrt}  
\bibliography{references}

\end{document}